\documentclass[10pt,twocolumn,letterpaper]{article}

\usepackage{cvpr}
\usepackage{graphicx}
\usepackage{amsmath}
\usepackage{amssymb}
\usepackage{paralist}
\usepackage{times}
\usepackage{multirow}
\usepackage{rotating}
\usepackage{url}
\usepackage{color}
\usepackage[symbol]{footmisc}

\usepackage[pagebackref=true,breaklinks=true,letterpaper=true,colorlinks,bookmarks=false]{hyperref}

\cvprfinalcopy % *** Uncomment this line for the final submission

\def\cvprPaperID{4} % *** Enter the CVPR Paper ID here

\ifcvprfinal
\fi

\begin{document}

% Example definitions.
% --------------------
%\def\x{{\mathbf x}}
%\def\L{{\cal L}}

% Title.
% ------
\title{Yoga-82: A New Dataset for Fine-grained Classification of Human Poses}
%
% Single address.
% ---------------
\author{Manisha Verma$^{1}$\thanks{Authors contributed equally to this work.}, Sudhakar Kumawat$^{2}$\footnotemark[1], Yuta Nakashima$^{1}$, Shanmuganathan Raman$^{2}$\\
 $^{1}$Osaka University, Japan $^{2}$Indian Institute of Technology Gandhinagar, India, \hspace{0.1em}\\
$^{1}$\{\tt\small mverma,n-yuta\}@ids.osaka-u.ac.jp $^{2}$\{\tt\small sudhakar.kumawat,shanmuga\}@iitgn.ac.in} 
% \author{Manisha Verma\\
% Osaka University\\
% Osaka, Japan\\
% {\tt\small mverma@ids.osaka-u.ac.jp}
% \and
% Sudhakar Kumawat\\
% IIT Gandhinagar\\
% Gandhinagar, India\\
% {\tt\small sudhakar@iitgn.ac.in}

% \and
% Yuta Nakashima\\
% Osaka University\\
% Osaka, Japan\\
% {\tt\small n-yuta@ids.osaka-u.ac.jp}

% \and
% Shanmuganathan Raman\\
% IIT Gandhinagar\\
% Gandhinagar, India\\
% {\tt\small shanmuga@iitgn.ac.in}
% }
\maketitle
\thispagestyle{empty}

%%%%%%%%% ABSTRACT

%Address and e-mail should NOT be added in the submission paper. They should be present only in the camera ready paper. 

%
\begin{abstract}
%Human pose recognition is a well-known problem in computer vision. Existing datasets for learning of poses are observed to be not challenging enough in terms of pose diversity, object occlusion and view points. This makes the pose annotation process relatively simple and restricts the application of the models that have been trained on them. To address this problem, we propose the concept of fine-grained hierarchical pose estimation and propose a new human pose classification dataset, Yoga-82,\footnote{The dataset will be available publicly after acceptance.} for large-scale yoga pose recognition with 82 classes. Yoga-82 consists of complex poses where fine annotations may not be possible. To resolve this, we provide hierarchical labels for yoga poses based on the body configuration of the pose. The dataset contains a three-level hierarchy including body positions, variations in body positions, and the actual pose names. We present the classification accuracy of the state-of-the-art convolutional neural network architectures on Yoga-82. We also present several hierarchical variants of DenseNet in order to utilize the hierarchical labels. 

Human pose estimation is a well-known problem in computer vision to locate joint positions. Existing datasets for learning of poses are observed to be not challenging enough in terms of pose diversity, object occlusion and view points. This makes the pose annotation process relatively simple and restricts the application of the models that have been trained on them. To handle more variety in human poses, we propose the concept of fine-grained hierarchical pose classification, in which we formulate the pose estimation as a classification task, and propose a dataset, Yoga-82\footnote[4]{ https://sites.google.com/view/yoga-82/home  }, for large-scale yoga pose recognition with 82 classes. Yoga-82 consists of complex poses where fine annotations may not be possible. To resolve this, we provide hierarchical labels for yoga poses based on the body configuration of the pose. The dataset contains a three-level hierarchy including body positions, variations in body positions, and the actual pose names. We present the classification accuracy of the state-of-the-art convolutional neural network architectures on Yoga-82. We also present several hierarchical variants of DenseNet in order to utilize the hierarchical labels.
\end{abstract}

\section{Introduction}
\begin{figure}[t]
    \centering
    \includegraphics[width=\columnwidth]{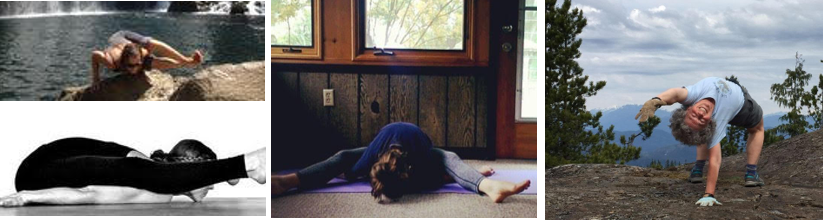}
    \caption{Example human poses from the yoga activity.}
    \label{Fig:vposes}
    %\vspace{-4.5mm}
\end{figure}

Human pose estimation has been an important problem in computer vision with its applications in visual surveillance \cite{chen2018shpd}, behaviour analysis \cite{holte2012human}, assisted living \cite{dias2020gaze}, and intelligent driver assistance systems \cite{martin2017real}. With the emergence of deep neural networks, pose estimation has achieved
drastic performance boost. To some extent, this success can be attributed to the availability of large-scale human pose datasets such as MPII \cite{andriluka20142d}, %MSCOCO \cite{lin2014microsoft},
FLIC \cite{sapp2013modec}, SHPD \cite{chen2018shpd}, and
LSP \cite{johnson2010clustered}. The quality of keypoint and skeleton annotations in these datasets play an important role in the success of the state-of-the-art pose estimation models. However, the manual annotation process is prone to human errors and can be severely affected by various factors such as resolution, occlusion, illumination, view point, and diversity of poses. For example, Fig.~\ref{Fig:vposes} showcases some human pose images for the yoga activity which inherently consist of some of the most diverse poses that a human body can perform. It can be noticed that some of these poses are too complex to be captured from a single point of view. This becomes more difficult with the changes in image resolution and occlusions. Due to these factors, producing fine pose annotations such as keypoints and skeleton for the target objects in these images may not be possible as it will lead to false and complex annotations.

\begin{figure}[b]
    \centering
    \includegraphics[width=\columnwidth]{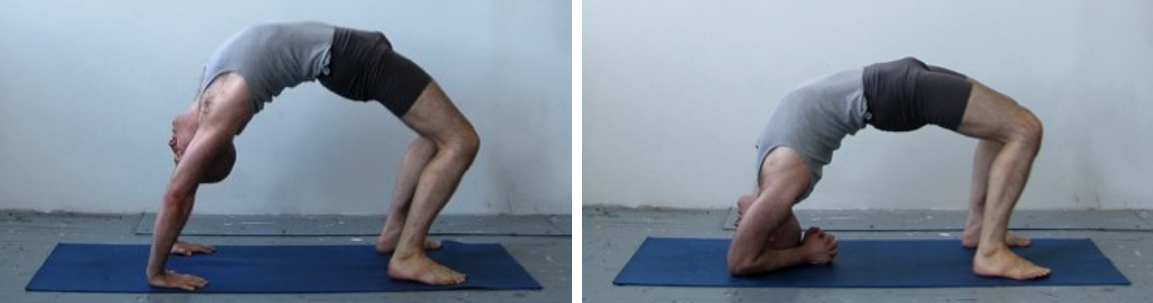}
    \caption{The \emph{upward bow pose} (left) and \emph{upward facing two-foot staff pose} (right). Both the poses have same superclass called \emph{up-facing wheel pose}.}
    \label{Fig:poses}
    %\vspace{-4.5mm}
\end{figure}

In order to solve this problem, we propose the concept of fine-grained hierarchical pose classification. Instead of producing fine keypoints and skeleton annotations for human subjects which may not be possible due to various factors, we propose hierarchical labeling of human poses where the classes are separated by the variations in body postures, which involve much in their appearances. One important benefit of hierarchical labeling is that the categorical error can be restricted to particular subcategories, such that it is more informative than the classic flat $N$-way classification. For example, consider the two yoga poses shown in Fig.~\ref{Fig:poses}, the \emph{upward bow pose} and the \emph{upward facing two-foot staff pose}. The two poses differ in the manner that the \emph{upward facing two-foot staff pose} puts headstand together with the \emph{upward bow pose}. Apart from this, both the poses have many similarities such as the way in which the back is bent (\emph{wheel pose}), the orientation of faces, and the placement of legs. Therefore, both these poses can be put in a single superclass pose called \emph{up-facing wheel pose}. An advantage of this type of label structure is that, once the network learns that it is a type of up-facing pose, it will not confuse it with any down-facing poses such as the \emph{cat-cow pose} which have a similar \emph{wheel} type structure. Further separation of the classes at the lowest level will help the network to focus on specific parts of the body. For example, the headstand part of the \emph{upward facing two-foot staff pose}.

\begin{figure*}[t]
    \centering
    \includegraphics[width=.99\textwidth]{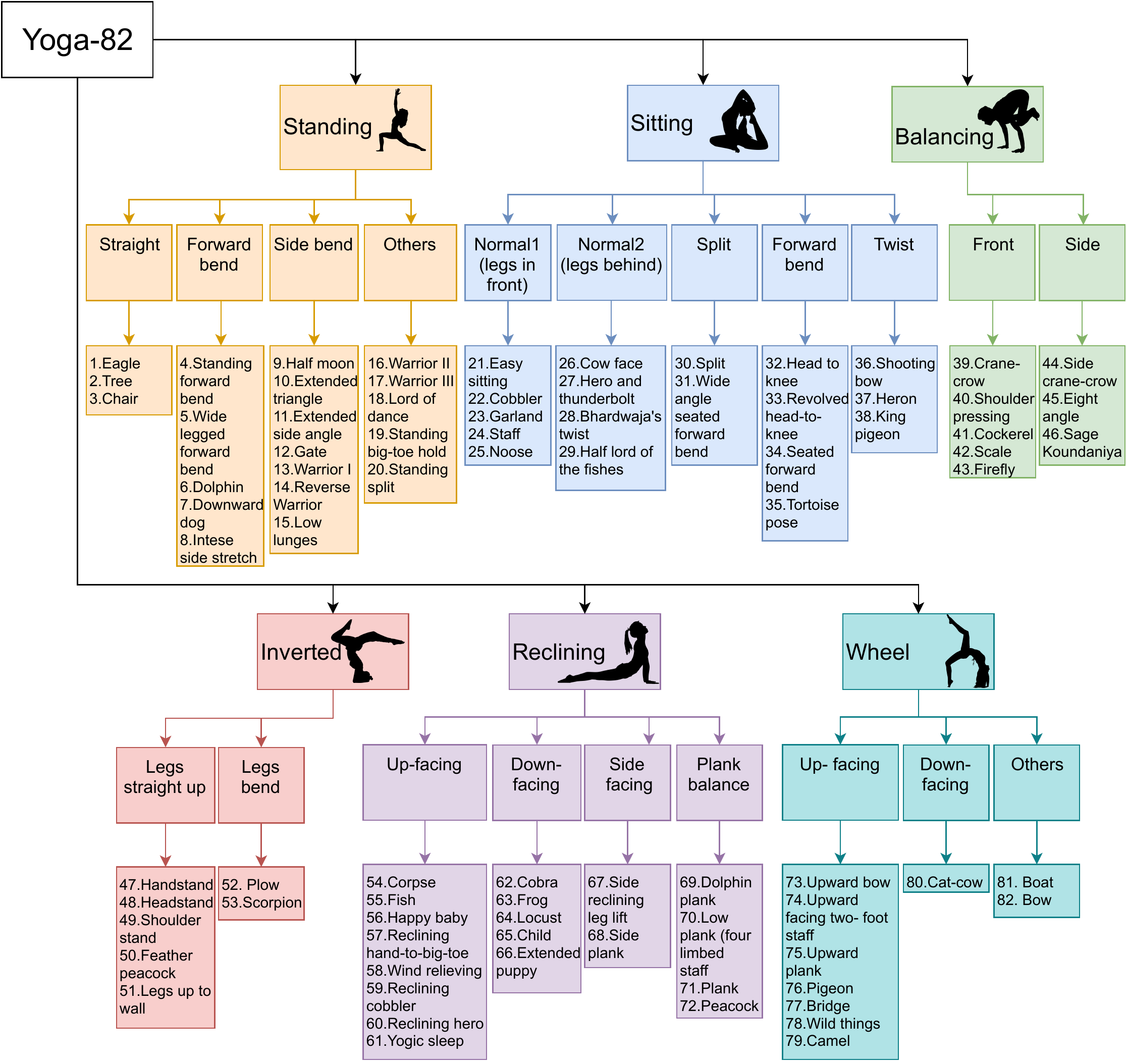}
    \caption{Yoga-82 dataset label structure. Hierarchical class names at level 1, 2, and 3.}
    \label{fig:yoga82}
\end{figure*}

In this work, building on the concept of fine-grained hierarchical pose classification (as discussed above), we propose a large-scale  yoga dataset. % that we call the Yoga-82 dataset.
We choose yoga activity since it contains a wide variety of finely varying complex body postures with rich hierarchical structures. 
%Although performing yoga is a process of transitioning from one body posture (normal) to another (yoga pose) and a sequence of images can define it better, the final posture defines the yoga pose and can be judged by a single image only.
This dataset contains over 28.4K yoga pose images distributed among 82 classes. These 82 classes are then merged/collapsed based on the similarities in body postures to form 20 superclasses, which are then further merged/collapsed to form 6 superclasses at the top level of the hierarchy (Fig.~\ref{fig:yoga82}). To the best of our knowledge, Yoga-82 is the first dataset that comes with class hierarchy information. In summary, the main contributions of this work are as follows.
\begin{itemize}
    \item We propose the concept of fine-grained hierarchical pose classification and propose a large-scale pose dataset called Yoga-82, comprising of multi-level class hierarchy based on the visual appearance of the pose.
    \item We present performance evaluation of pose recognition on our dataset using well-known CNN architectures.
    \item We present modifications of DenseNet in order to utilize the hierarchy of our dataset for achieving better pose recognition.
\end{itemize}

\begin{table}[!t]
%\footnotesize
\caption{Comparison of human pose datasets.}
\footnotesize
\centering
\begin{tabular}{lllll}
\hline
Datasets & \#Train & \#Test & Source     & Target poses          \\ \hline
MPII \cite{andriluka20142d}     & 25,000     & -      & YouTube      & Diverse \\
%MSCOCO \cite{lin2014microsoft}  & 118k    & 41k    & Flickr        & diverse            \\
LSP \cite{johnson2010clustered}     & 1,000    & 1,000   & Flickr       & Sports  \\
LSP-Ext. \cite{johnson2011learning} & 1,0000 & - & Flickr & Sports \\
FLIC  \cite{sapp2013modec}     & 6,543    & 1,016   & Movies       & Diverse            \\

SHPD \cite{chen2018shpd}    & 18,334   & 5,000   & Surveillance & Pedestrian  \\ 
Yoga-82 & 21,009   & 7,469 & Bing & Yoga \\ 
\hline      
\end{tabular}
\label{tab:datsetsold}
\end{table}

\noindent{\textbf{Related work.}}
Human pose estimation has been an important problem in computer vision and many benchmark datasets have been proposed in the past. Summary of some of the most commonly used human pose datasets is presented in Table~\ref{tab:datsetsold}. Many of these datasets are collected from the sources such as online videos, movies, images, sports videos, etc. 
Some of them provide rich label information but lack in human pose diversity. Most of the poses in these datasets (\cite{andriluka20142d}, \cite{chen2018shpd}, and \cite{sapp2013modec}) are of standing, walking, bending, sitting, etc. and not even close to comparison with complex yoga poses (Fig.~\ref{Fig:vposes}). Chen \etal \cite{chen2018shpd}, recently observed that 
%most of these datasets lack in two aspects. First, the diversity of the human poses in these datasets is very limited. Second, 
the images considered for annotations are of very high quality with large target objects. For example, 
%in MSCOCO \cite{lin2014microsoft}, most of the human poses are upright poses and diverse poses such as bending, lying, and other complex poses are very less; 
in the MPII dataset \cite{andriluka20142d}, around 70\% of the images consists of human objects with height over 250 pixels. Thus, without much diversity in human poses and target object size in these datasets, they can not meet the high-quality requirements of applications such as behaviour analysis. Our proposed Yoga-82 dataset is very different from these datasets in the two aspects discussed above. We choose yoga activity, which we believe consists of some of the most diverse and complex examples of human poses. Furthermore, the images considered from the wild are with different viewpoints, illumination conditions, resolution, and occlusions. Few works have been done on yoga pose classification for applications such as self training
\cite{chen2014yoga,yadav2019real,qiao2017real,islam2017yoga}. However these works involve yoga dataset with a less number of images or videos and does not consider vast variety of poses. Hence, they lack in generalization and are far from complex yoga pose classification. 

\begin{figure}[t]
    \centering
    \includegraphics[width =.17\textwidth]{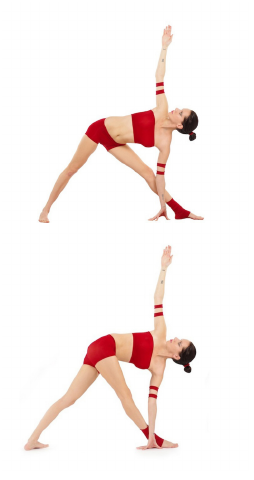} 
    \includegraphics[width =.08\textwidth]{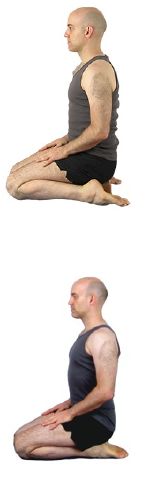} 
    \includegraphics[width =.16\textwidth]{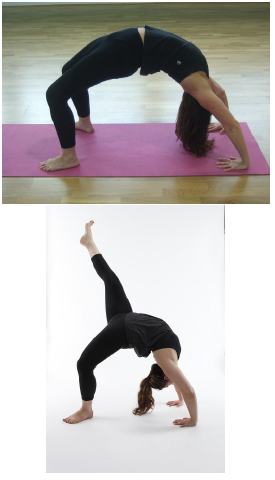}\\
    \vspace{-1mm}
    
    (a)  \hspace{17mm}  (b) \hspace{17mm}  (c)\\
    \vspace{1mm}
    \includegraphics[width =.17\textwidth]{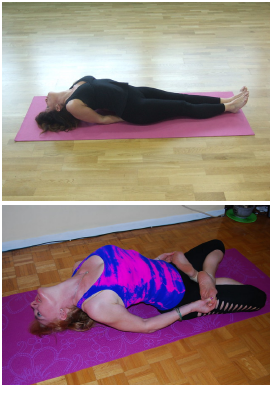}
    \includegraphics[width =.16\textwidth]{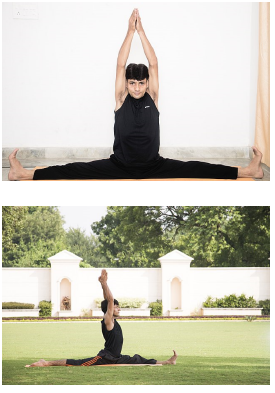}\\
    \vspace{-1mm}
    (d) \hspace{26mm} (e)\\
    \caption{Different variations of the same pose in one class (a) \textit{Extended triangle pose} and \textit{revolved triangle pose}, (b) \textit{Hero pose} and \textit{thunderbolt pose}, (c) \textit{Upward bow pose} and its variation, (d) \textit{Fish pose} and its variation, and (e) \textit{Side spilt} and \textit{front split pose}.}
    \label{fig:samepose}
\end{figure}
% \begin{figure}[t]
%     \centering
%     \includegraphics[width =.14\textwidth]{similar_variations_in_one_class_hori_11.pdf} 
%     \includegraphics[width =.07\textwidth]{similar_variations_in_one_class_hori_12.pdf} 
%     \includegraphics[width =.12\textwidth]{similar_variations_in_one_class_hori_13.pdf}
%     %\vspace{-1mm}
%     \includegraphics[width =.13\textwidth]{similar_variations_in_one_class_hori_14.pdf}

%     (a)  \hspace{14mm}  (b) \hspace{12mm}  (c)  \hspace{16mm}  (d)\\
%     %\vspace{1mm}
%         %\includegraphics[width %=.16\textwidth]{similar_variations_in_one_class_hori_15.pdf}
%     %\\
%     %\vspace{-1mm}
%     %(d) \hspace{26mm} (e)\\
%     \caption{Different variations of the same pose in one class.% (a) \textit{Extended triangle pose} and \textit{revolved triangle pose}, (b) \textit{Hero pose} and \textit{thunderbolt pose}, (c) \textit{Upward bow pose} and its variation, and (d) \textit{Fish pose} and its variation.}%, and (e) \textit{Side spilt} and \textit{front split pose}.
%     }
%    \label{fig:samepose}
%\end{figure}
\begin{figure}[!h]
	\centering
	\includegraphics[width=.49\textwidth]{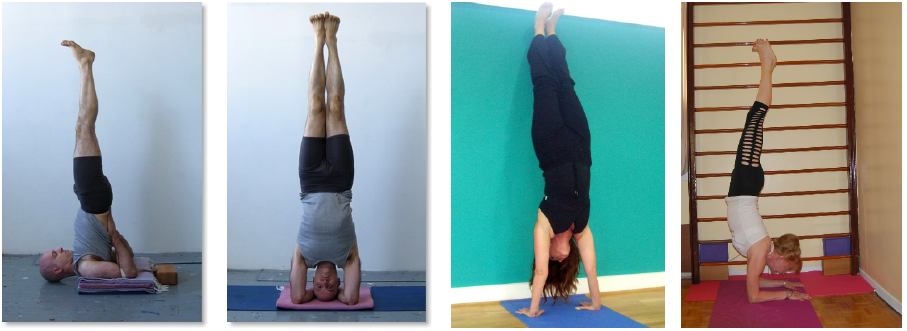}\\
	(a) \textit{Inverted poses.}\\
	\includegraphics[width=.49\textwidth]{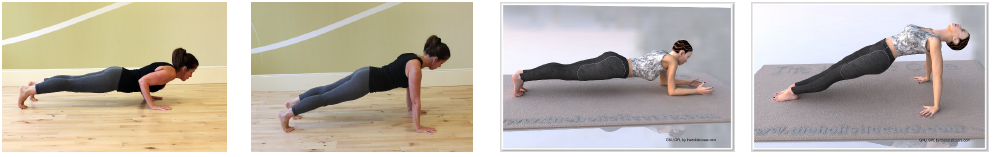}\\
	(b) \textit{Plank poses.}\\
	\includegraphics[width=.45\textwidth]{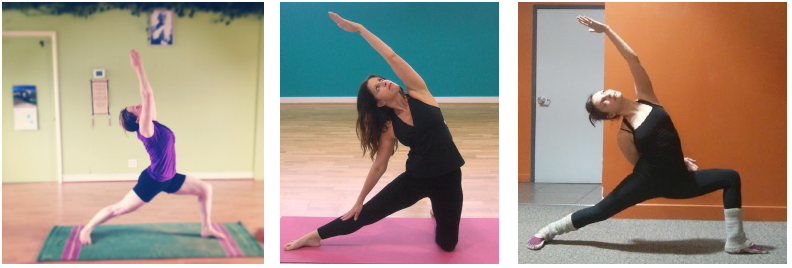}\\
	(c) Standing poses\\
	\includegraphics[width=.49\textwidth]{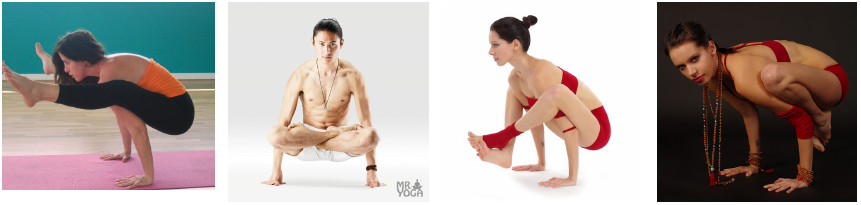}\\
	(d) Balancing poses\\
	\caption{Some example of different classes in very similar appearances.}
	\label{fig:diffpose}
\end{figure}
%\vspace{-4mm}
\section{The Yoga-82 Dataset}

% We propose a large-scale yoga dataset with wide variety of yoga poses. 
% %The dataset is comprised of images containing a person/people in a certain yoga pose. 
% This section discusses about data collection, hierarchical labelling, and analysis over the dataset.  
\noindent\textbf{Data Acquisition.}
The dataset contains yoga pose images downloaded from web using the Bing search engine. The taxonomy about yoga poses (name and appearance) is collected from various websites and books \cite{iyengar1965light, kaminoff2011yoga, yogajournal, wikiyoga}. Both Sanskrit and English names of yoga poses were used to search for images and the downloaded images were cleaned and annotated manually. Every image contains one or more people doing the same yoga pose. Furthermore, images have poses captured from different camera view angles. There are a total of 82 yoga pose classes in the dataset. The dataset has a varying number of images in each class from 64 (min.) to 1133 (max.) with an average of 347 images per class. Some of the images are downloaded from a specific yoga website. Hence, they contain only yoga pose with clean background. However, there are many images with random backgrounds (e.g., forest, beach, indoor, etc.). Some images only contain silhouette, sketch, and drawing version of yoga poses and they were kept in the dataset as yoga pose is more about the overall structure of body and not the texture of clothes and skin.
%We have not cropped the images and retained as they were being downloaded from the web. 
For the sake of easiness in understanding and readability, here we use only English names for the yoga poses. However, Sanskrit names are available as well in the dataset for reference.

%both yoga classes 
%and images
%cleaning

\noindent\textbf{Label Hierarchy and Annotation.}
%Each yoga pose
%Hierarchy
% \begin{figure}
%     \centering
%     \includegraphics[width=.32\textwidth]{similar_variations_in_one_class11.pdf}\\
%     (a) Extended triangle pose and revolved triangle pose\\
%     \includegraphics[width=.29\textwidth]{similar_variations_in_one_class12.pdf}\\
%     (b) Hero pose and thunderbolt pose\\
%     \includegraphics[width=.35\textwidth]{similar_variations_in_one_class13.pdf}\\
%     (c) Fish pose and fish pose variation\\
%     \includegraphics[width=.35\textwidth]{similar_variations_in_one_class14.pdf}\\
%     (d) Upward bow pose and upward bow pose variation\\
%     \includegraphics[width=.35\textwidth]{similar_variations_in_one_class15.pdf}\\
%     (e) Side spilt and front split pose\\
%     \caption{Different variations of same pose in one class [image citations]}
%     \label{fig:samepose}
% \end{figure}
Existing pose datasets  (Table~\ref{tab:datsetsold}) available publicly for evaluation do not impose hierarchical annotations. Hierarchical annotations can be beneficial for part-based learning in which few parts of the network will learn features based on the hierarchical classes. Many hierarchical networks have been observed to perform better as compared to their baseline CNN models \cite{zhu2017b, fu2018cnn}. Hierarchical annotations are  beneficial for learning the network as they provide rich information to users not only about the pose names but also about the body postures (\textit{standing}, \textit{sitting}, etc.), the effect on the spine (e.g., \textit{forward bend}, \textit{back bend} in \textit{wheel pose}, etc.), and others (e.g. \textit{down-facing} or \textit{up-facing}).

Our labels are with a three-level hierarchical structure where the third level is a leaf node (yoga pose class). There are 6, 20, and 82 classes in the first-, second-, and third-levels, respectively as illustrated in Fig.~\ref{fig:yoga82}. References for these classes have been collected from websites and books \cite{singleton2010yoga, iyengar1965light, kaminoff2011yoga, yogajournal, wikiyoga, tummee}.
There is no established hierarchy in yoga poses. However, \textit{standing}, \textit{sitting}, \textit{inverted}, etc. are well defined as per their configuration. In this work, we have taken guidelines from \cite{singleton2010yoga, iyengar1965light, kaminoff2011yoga, yogajournal, wikiyoga} in order to define the first level classes and defined a new class (\textit{wheel}) based on the posture of the subject's body in a certain pose. The second level further divides the first level classes into different classes as per subject's body parts configuration. 
However, it is hard to define 82 leaf classes in 6 super classes perfectly. Yet, we have made an attempt to briefly describe the 6 first level classes as follows.

%\textbf{Class level 1:} \\
\noindent\textbf{\textit{Standing:}} Subject is standing while keeping their body straight or bending. Both or one leg will be on the ground. When only one leg is on the ground, the other leg is in air either held by one hand or free.

\noindent\textbf{\textit{Sitting:}} Subject is sitting on the ground. Subject's hip will be on ground or very close to the ground (e.g., \textit{garland pose}). 

\noindent\textbf{\textit{Balancing:}} Subject is balancing their body on palms. Both the palms are on ground and the rest of the body is in air. Subject's body is not in the inverted position.

\noindent\textbf{\textit{Inverted:}} Subject's body is upside down. Lower body is either in air or close to the ground (e.g., \textit{plow pose}).

\noindent\textbf{\textit{Reclining:}} Subject's body is lying on the ground. Either spine (\textit{upward facing}) or stomach (\textit{downward facing}) or side body (\textit{side facing}) touching or very near to the ground, or subject's body is in $ 180^ {\circ} $ angle (approximately) alongside ground (e.g, \textit{plank poses}). 

\noindent\textbf{\textit{Wheel:}} Subject’s body is in half circle or close to half circle on ground. In \textit{upward facing} or \textit{downward facing poses}, both the palm and the feet will touch the ground. In \textit{others} category, either only hip or stomach will touch the ground; the other body parts will be in air.

All the class levels are shown in Fig.~\ref{fig:yoga82}. The class names and their images are based on the body configuration. For example, \textit{forward bend} is a second-level class in both standing and sitting. As it is clear from its name, this class includes poses where the subject needs to bend forward while standing or sitting. Similar names are given to the other second-level classes. The poses that do not fit in any second-level classes are kept in \textit{others} (\textit{standing}), \textit{twist} (\textit{sitting}), \textit{normal1} (\textit{sitting}), \textit{normal2} (\textit{sitting}), etc.

\setlength{\tabcolsep}{8pt}
\begin{table*}[!htb]
%\footnotesize
\centering
\caption{Performance of the state-of-the-art architectures on Yoga-82 using third-level class (82 classes). }
\label{tab:benchmark}
\begin{tabular}{lrrrll} 
\hline
 \textbf{Architecture}  & \textbf{Depth}  & \textbf{\# Params}  & \textbf{Model size} &    \textbf{Top-1}   & \textbf{Top-5}            \\ 
\hline
ResNet-50                                & 50                               & 23.70 M                              & 190.4 MB                             &  63.44           & 82.55                     \\ 

ResNet-101                               & 101                              & 42.72 M                              & 343.4 MB                             &  65.84           &  84.21                    \\ 

ResNet-50-V2                             & 50                               & 23.68 M                              & 190.3 MB                             &  62.56           &  82.28                    \\ 

ResNet-101-V2                            & 101                              & 42.69 M                              & 343.1 MB                             &  61.81           &  82.39                    \\ 

%Inception-V3                             & 159                              & 21.93 M                              & 176.6 Mb                             &  63.20           &  84.19                    \\ 

DenseNet-121                             & 121                              & 7.03 M                               & 57.9 MB                              &  73.48           &  90.71                    \\ 

DenseNet-169                             & 169                              & 12.6 M                               & 103.4 MB                             & 74.73            &  \textbf{91.44}                    \\ 

DenseNet-201                             & 201                              & 18.25 M                              & 149.1 MB                             &  \textbf{74.91}  &  91.30                    \\ 

MobileNet                                & 88                               & 3.29 M                               & 26.7 MB                              &  67.55           &  86.81                    \\ 

MobileNet-V2                             & 88                               & 2.33 M                               & 19.3 MB                              &  71.11           &  88.50                    \\ 

ResNext-50                               & 50                               & 23.15 M                              & 186.1 MB                             &  68.45           &  86.42                    \\ 

ResNext-101                              & 101                              & 42.29 M                              & 340.2 MB                             & 65.24            &  84.76                    \\
\hline
\end{tabular}
\end{table*}

\begin{table*}
%\footnotesize
\centering
\caption{Classification performances of our three variants. L1, L2, and L3 stand for the first-, second-, and third-level classification, respectively.}
\label{tab:PM}
\begin{tabular}{llllllll} 
\hline
\multirow{2}{*}{\textbf{Network} } & \multirow{2}{*}{\textbf{\# Params} } & \multicolumn{3}{l}{\textbf{Top-1} } & \multicolumn{3}{l}{\textbf{Top-5} }  \\ 
\cline{3-8}
                                   &                                      & L1    & L2    & L3                  & L1    & L2    & L3                   \\ 
\hline
Variant 1                          & 18.27 M                              & 83.84 & 85.10 & 79.35               & 99.40 & 97.08 & 93.47                \\ 

Variant 2                          & 18.27 M                              & 89.81 & 84.59 & 79.08               & 99.83 & 97.03 & 92.84                \\ 

Variant 3                          & 22.59 M                              & 87.20 & 84.42 & 78.88               & 99.69 & 97.28 & 92.66                \\
\hline
\end{tabular}
\end{table*}
%\end{group}
\noindent\textbf{Analysis over our Dataset.}
Some of the poses have variations of their own. For example, \textit{extended triangle pose} and \textit{revolved triangle pose}, \textit{head-to-knee pose} and \textit{revolved head-to-knee pose}, \textit{hero pose}, \textit{reclining hero pose}, etc. 
These poses are kept in the same class or different classes in the third level based on the differences in their visual appearances.
For example, \textit{extended triangle pose} and \textit{revolved triangle pose} (Fig.~\ref{fig:samepose}(a)) are in the same class, while \textit{head-to-knee pose} and \textit{revolved head-to-knee pose} are in different classes.

Some completely different poses (e.g. \textit{hero pose} and \textit{thunderbolt pose}, \textit{side spilt pose} and \textit{front split pose}, etc.) 
are kept in same third-level classes as they appear to be very similar to each other. For example, \textit{hero pose} and \textit{thunderbolt pose} (Fig.~\ref{fig:samepose}(b)) have a minute difference that legs to be placed near the thighs and under the thighs, respectively. Other class separations were carefully made using suggestions of three of the authors based on the appearance of the poses.

Our dataset is very challenging in terms of similarity between different classes. There are many classes at the third level that are very similar to each other that are treated as different poses. For example, \textit{inverted poses} (level 2) has poses that differ from each other if the subject is in inverted position and balancing their body up straight on hands (\textit{handstand pose}), head (\textit{headstand pose}), or forearms (\textit{feathered peacock pose}) as shown in Fig.~\ref{fig:diffpose}(a). Similarly, \textit{plank poses} differ from each other based on plank's height from the ground and whether its on palms or forearms (Fig.~\ref{fig:diffpose}(b)). 
Few similar poses are shown in Fig.~\ref{fig:diffpose}. These poses make the dataset very challenging as this is not covered in any previous pose datasets \cite{chen2014yoga}.  
%\vspace{-4mm}

\section{Experiments}
%Experiments on existing state-of-the-art methods
%Evaluation of the proposed network and comparison
%Evaluation on CIFAR 100
We divide our experiments into two parts. In the first part, we conduct benchmark experiments on the Yoga-82 dataset. In the second part, we present three CNN architectures that exploit the class hierarchy in our Yoga-82 dataset to analyze the performance using hierarchical labels. 
%%%%%%%%%%%%%%%%%%%%%%%%%%%%%%%%%%%%%%%%%%%

\subsection{Benchmarking Yoga-82 Dataset}
We evaluate the performance of several popular CNN architectures on the Yoga-82 dataset that have recently achieved state-of-the-art accuracies on image recognition tasks on the  ImageNet \cite{deng2009imagenet} dataset.

\noindent\textbf{Benchmark models.} Table~\ref{tab:benchmark} gives a comprehensive list of  network architectures that we used for benchmarking our Yoga-82 dataset. They are selected such that they differ in  structures, depth, convolutional techniques, as well as computation and memory efficiencies. For example, ResNet \cite{he2016deep,he2016identity} and DenseNet \cite{huang2017densely} differ in the manner the skip connections are applied. 
%Inception-V3 \cite{szegedy2016rethinking} uses factorized convolutions.
MobileNet \cite{howard2017mobilenets,sandler2018mobilenetv2} uses separable convolutions for better computational and memory efficiency. ResNext \cite{xie2017aggregated} uses group convolutions for better performance and reduces space-time complexity. 

\noindent\textbf{Experimental protocol and setting.} All our experiments were conducted on a system with Intel Xeon Gold CPU (3.60 GHz $\times$ 12), 96 GB RAM, and an NVIDIA Quadro RTX 8000 GPU with 48 GB memory. We used Keras with Tensorflow backend as the deep learning framework. For training the networks, we used stochastic gradient descent (SGD) with momentum 0.9. We started with a learning rate of 0.003 and decreased it by the factor of 10 when the validation loss plateaus. All weights were initialized with the orthogonal initializer. 
%Weight decay is used with the weight of $5 \times 10^{-4}$??.
We did not apply any data augmentation techniques on the input images. All images were resized to $224\times224$, before feeding into the networks. %Both pre-trained and non pre-trained models were used. For pre-trained models, we used the weights provided in Keras trained on ImageNet.
We split our dataset into training and testing sets, which contain 21009 and 7469 images, respectively. As mentioned earlier, we provide train-test splits of the dataset for consistent evaluation and fair comparison over the dataset in future.  

 \begin{figure*}[htb]
     \centering
     \includegraphics[width=.99\textwidth]{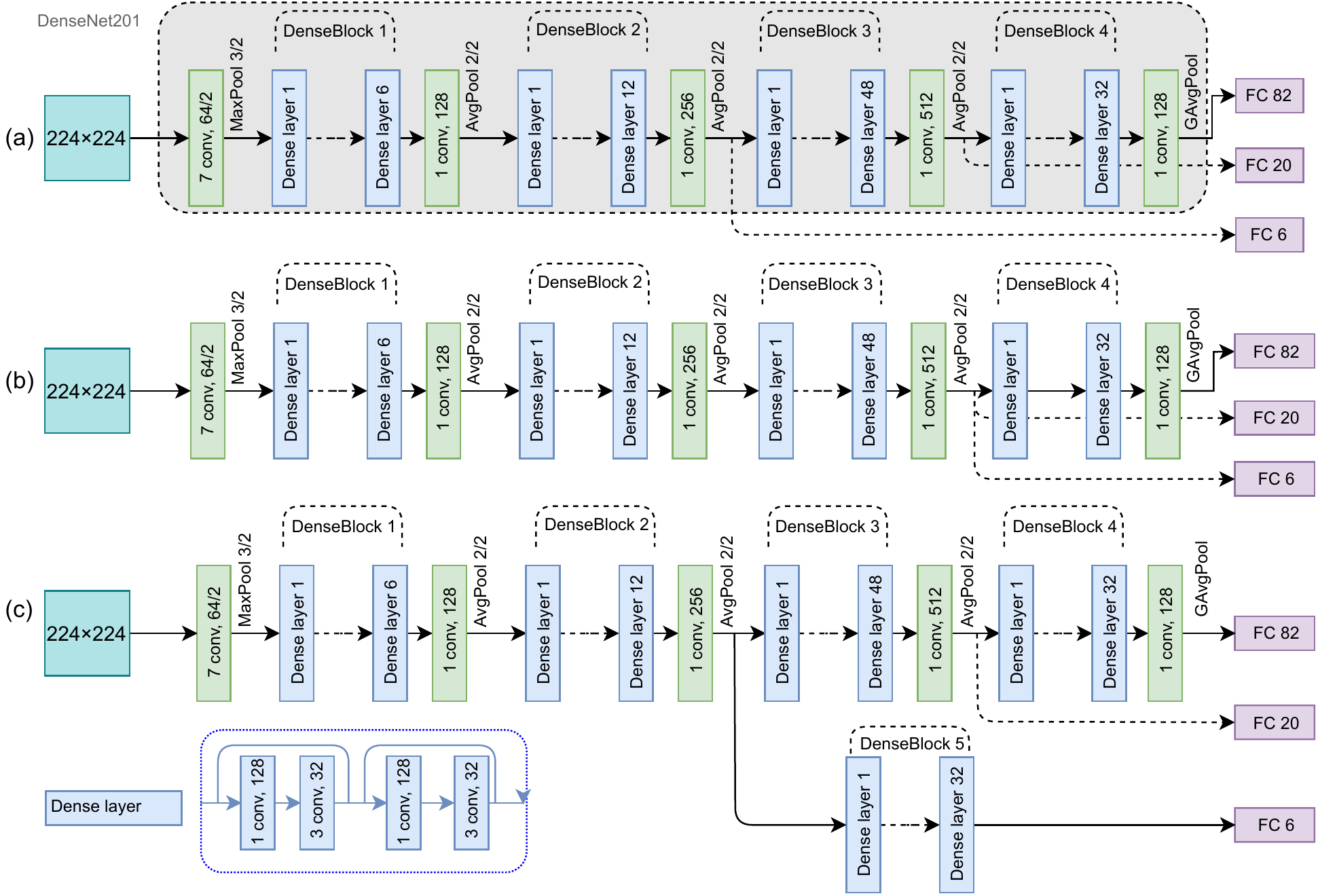}
     \caption{DenseNet-201 modified hierarchical architectures.}
     \label{fig:pm}
 \end{figure*}
\noindent\textbf{Results.} The results of the benchmark experiments are shown in Table~\ref{tab:benchmark}. Both the top-1 and top-5 classification accuracies are reported. We observe that deeper networks have a clear edge over their shallower versions. For example, 101-layer ResNet architecture (ResNet-101) outperformed its 50-layer variant (ResNet-50). 
%DenseNet and ResNext behaved similarly.
Furthermore, deeper networks with dense skip connections, such as the DenseNet architectures, performed better than the networks with sparse skip connections. %When trained from scratch on the Yoga-82 dataset,
DenseNet-201 gives the best performance, achieving top-1 classification accuracies of 74.91\% .
%When we fine-tune the pre-trained model, ResNext-101 achieved the best top-1 classification accuracies of 92.74\% and 92.97\% on the validation and test sets, respectively. This is closely followed by DenseNet-201, which achieved the accuracies of 92.08\% and 92.09\%.

\subsection{Hierarchical Architectures}
\label{sec:arch}
Our dataset, Yoga-82, provides a rich hierarchical structure in the labels, which can be utilized in order to enhance the performance of pose recognition.
Based on \cite{zhu2017b}, 
we modify Densenet-201 architecture to make use of the structure. That is, due to the hierarchical structure, label prediction in any level can be deducted from the third-level prediction results. However, since the hierarchy is based much on visual similarity between different poses, training with upper-level labels may help lower-level boost the prediction in lower-level label, and vice versa.
%Therefore, we add branches networks for each level and corresponding losses. We employ DenseNet-201 as our base architecture and modify it in three different ways.

%\noindent \textbf{Protocol 1:} In first protocol, full architecture of DenseNet201 is utilized to classify all three class levels. Each class level is classified using final feature vector of Densenet201 as shown in Fig. \ref{fig:pm}(a). This architecture only takes advantage of hierarchy at final layer and all fine-to-coarse classes have access of whole network. Global average pooling is taken over final feature map and resultant vector is connected to three classification layers of 6 ,20, and 82 classes for class level 1,2, and 3 respectively. Categorical-cross entropy loss is computed for all three labels and weighted sum is evaluated for final loss as follows. 
%\begin{equation}
%\footnotesize
%    L = w_1\sum_{i=1}^{c_1}C1_i log(C1_{i}^{'}) + w_2\sum_{j=1}^{c_2}C2_j log(C2_{j}^{'}) + w_3\sum_{k=1}^{c_3}C3_k log(C3_{k}^{'}) 
%\end{equation}
%\noindent where $c_1$, $c_2$, and $c_3$ are 6, 20, and 82 respectively. $C1$, $C2$, and $C3$ are ground truth labels of class level 1, level 2., and level 3 respectively. Similarly, $C1^{'}$, $C2^{'}$, and $C3^{'}$ are predicted probabilities respectively. In this network, all three weights are set to 1 as all three labels are important and all three classification layers have access to whole network. 

\noindent\textbf{Variant 1.} In this variant, hierarchical connections are added in DenseNet-201 after DenseBlock 2 and DenseBlock 3 for class level 1 (6 classes) and class level 2 (20 classes), respectively, as shown in Fig.~\ref{fig:pm}(a). Coarser classes are classified at the middle layers and finer classes are at the end layers of the network. The intuition behind this variant is to utilize hierarchy structure in the dataset. Initial-to-mid layers learn to classify the first level and the details in the input image is passed on to next layers for the second-level classification, and so on. Layers shared by all three levels (up to DenseBlock 2) learn basic structure of pose and further layers refine it for specific details.
The branch for the first-level classification applies batch normalization and the ReLU activation, followed by global average pooling. The same applies to the branch for the second-level classification. The main branch is for the third-level classification with 82 classes. 
Softmax-cross entropy loss is computed for all three levels and weighted sum is evaluated as the final loss as follows: 
%\vspace{-1mm}
\begin{equation}
\footnotesize
   L =\sum_{i=1}^{3} w_i\sum_{j=1}^{N_i} t_{ij} \log(y_{ij}),
   % L = w_1\sum_{i=1}^{c_1}C1_i log(C1_{i}^{'}) + w_2\sum_{j=1}^{c_2}C2_j log(C2_{j}^{'}) + w_3\sum_{k=1}^{c_3}C3_k log(C3_{k}^{'}) 
\end{equation}
%\vspace{-1mm}

\begin{figure*}[htb]
    \centering
    \includegraphics[width=.95\textwidth]{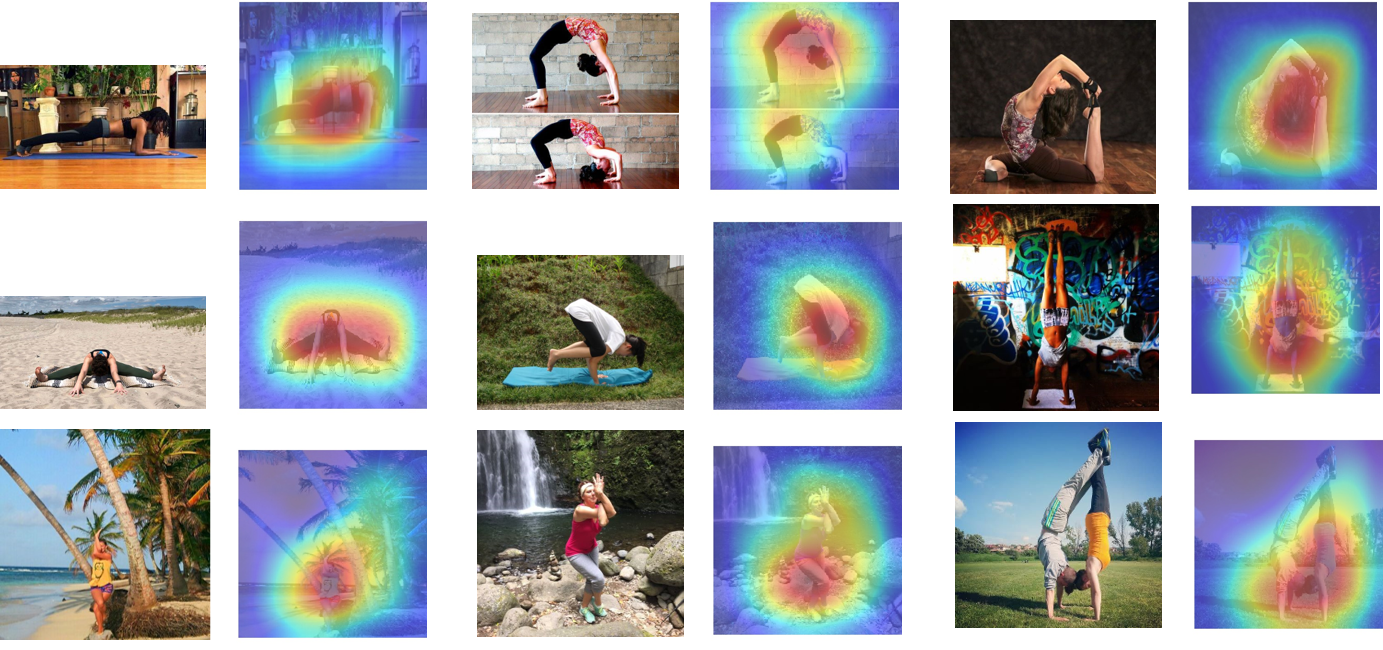}
    \caption{Activation maps learned using variant 2.}
    \label{fig:actimap}
    %\vspace{-.1in}
\end{figure*}
\noindent where $N_i$ ($i=1,2,3$) is the number of labels in level $l$, i.e., 6, 20, and 82 for $i =$ 1, 2, and 3, respectively. $t_{ij} \in \{0, 1\}$ is ground truth for label $j$ of level $l$. $y_{ij}$ is the output of the softmax layer. $w_i$ is the weight for level $i$. All weights are set to one as we consider that all levels are equally important.  

\noindent\textbf{Variant 2.} In variant 1, the first-level classifier does have access to only DenseBlock 1 and DenseBlock 2 that comprises of 6 and 12 dense layers, respectively, whereas DenseBlock 3, which classifies level 2, has 48 dense layers. Hence, the accuracy of the first-level classifier may be degraded in variant 1 because of insufficient representation capability. 
Since our focus is to classify images into classes in all three levels correctly, we make branches for the first- and second-level classifiers from the same position (Fig.~\ref{fig:pm}), so that the first-level classifier can have more representation capability.
We classify both levels after DenseBlock 3 as illustrated in Fig.~\ref{fig:pm}(b). Batch normalization, the ReLU activation, global average pooling, and loss function are the same as variant 1.

\noindent\textbf{Variant 3. } Another attempt to classify all three levels equally is made in variant 3.
We employ a similar architecture as variant 1, except that we add DenseBlock 5 with 32 dense layers for the first-level classifier branch (Fig.~\ref{fig:pm}(c)). This variant gives more trainable parameters to the first-level classifiers while keeping the hierarchical structure of network. This variant increases the number of parameters compared to the others.

\noindent\textbf{Results and Discussion.}
The performances of all three variants are presented in Table \ref{tab:PM} along with the numbers of parameters. All three variants stem from DenseNet-201 and thus the numbers of parameters differ only because of the addition of DenseBlock 5 in variant 3. 
%The reason behind these three specific structures is to analyse hierarchical training of the network.
Clearly, the hierarchical structures boosted the performances. As shown in Tables \ref{tab:benchmark} and \ref{tab:PM}, the accuracy of the third-level classifier (82 classes) was boosted from 74.91\% to 79.35\% with hierarchical connections added in DenseNet-201. %The accuracy of the pre-trained networks slightly improved as well compared to the original DenseNet-201. %Since, we did not pre-train our networks from scratch using hierarchy labels of Imagenet, might be the reason behind this minor improvement.  

We can see that the third-level classifiers (L3 in Table \ref{tab:PM}) give similar accuracies varying within 1\% in all three variants. In contrast, we see huge variations in the accuracy of the first-level classifier. 
%As mentioned in Section \ref{sec:arch}, variant 1 has less layers responsible for the first-level classification than the other two variants.
From these results, we may say that the performance depends more on the number of layers or the parameters responsible for a certain level classifier as well as on the number of classes. For example, variant 1 uses two DenseBlocks ($6+12$ dense layers) for the first-level classification and three DenseBlocks ($6+12+48$ dense layers) for the second-level classification. This huge gap between the numbers of parameters used for the first- and second-level classification may cause the difference in performances. This gap is reduced with variant 2 whose first- and second-level classifiers branch at the same point  (i.e., after DenseBlock 3).  As expected, the accuracy of the first level is less than that of the second level 2 and the accuracy of the second level is less than that of the third level. % in both with and without pre-training. 
Similarly, variant 3 has extra layers added for the first-level classification.  Hence, the accuracies decrease in the order of the first level to the third level classifiers. 
%when pre-trained weights are not used. Since our newly added layers were not pre-trained on ImageNet, the performance is not as good as variant 2 but is better than variant 1 with pre-training. 
Variant 3 increases the performance at the cost of additional parameters in the network. In conclusion, variant 2 can balance well.  

In Fig.~\ref{fig:actimap}, we present the class activation maps obtained from variant 2 using \cite{zhou2016learning}.
%for some images selected from our validation set. Note that these selected images have cluttered background and may contain multiple people in a single image. 
It can be observed that our model responded to the person doing a certain pose. Furthermore, we observe that, for a particular pose, the model focuses on one or specific parts of the body. For example,  for \textit{eagle pose} (Fig.~\ref{fig:actimap}, second column), the model focused on the configuration of the legs of the person.
%\vspace{-4mm}
\section{Conclusion}
In this work, we explored human pose recognition from a different direction by proposing a new dataset, Yoga-82, with 82 yoga pose classes. We define a hierarchy in labels by grasping the knowledge of body configurations in yoga poses. In particular, we present a three-level hierarchical label structure consisting of 6, 20, and 82 classes in the first to third levels. We conducted extensive experiments using popular state-of-the-art CNN architectures and reported benchmark results for the Yoga-82 dataset. We present modified DenseNet architecture to utilize the hierarchy labels and get a performance boost as compared to the flat $n$ label classification. It is evident that hierarchy information provided with dataset improves the performance because of additional learning supervision. It is visible from results that there is sufficient room for accuracy improvement in yoga pose classification. In future, we will focus on adding explicit constraints among predicted labels for different class levels.
%\vspace{-4mm}
%\subsection{Ablation study}
%
%Evaluation of the proposed method on different level of hierarchy and (comparison with state-of-the-art)
%
%the proposed method evaluation from root to leaf, root to leaf including one sub level, root to leaf including 2 sub levels.


% \bibliographystyle{IEEEbib}
% \bibliography{egbib}

\end{document}